\documentclass{article}

\usepackage[sort]{natbib}
\usepackage[preprint]{neurips_2026}

\usepackage[utf8]{inputenc}
\usepackage[T1]{fontenc}
\usepackage{hyperref}
\usepackage{url}
\usepackage{booktabs}
\usepackage{amsfonts}
\usepackage{nicefrac}
\usepackage{microtype}
\usepackage{xcolor}
\usepackage{graphicx}
\usepackage{amsmath, amsthm, amssymb}
\usepackage{thmtools}
\usepackage{setspace}
\usepackage{float}
\usepackage{framed}
\usepackage{tcolorbox}
\usepackage{bbm}
\usepackage{caption}
\usepackage{subcaption}
\usepackage[ruled,vlined,lined,commentsnumbered,linesnumbered]{algorithm2e}
\usepackage{algorithmic}
\usepackage{dsfont}
\usepackage{multirow}
\usepackage{makecell}
\usepackage{adjustbox}
\usepackage{wrapfig}

\title{DSA: Density-aware Sample-specific Attack}

\author{%
  Qiyuan Wang
  \\ Texas A\&M University \\
  \And
  Yao Li\\
  University of North Carolina at Chapel Hill \\
  \And
  Raymond K.~W.~Wong \\
  Texas A\&M University \\
}

\begin{document}

\maketitle

\begin{abstract}
Despite recent progress in backdoor attacks, existing methods remain susceptible to post-training defenses that erase the backdoor through fine-tuning or pruning. We revisit the core objectives of backdoor attacks and derive principled criteria characterizing optimal sample-specific trigger construction under a Bayes-optimal model of the victim's training. Our analysis reveals that both attack success and clean-accuracy preservation are simultaneously optimized when triggered samples are steered into low-density regions of the clean data distribution, a distributional condition that controls all moments of the poisoned distribution at once rather than a handful of input-space summary statistics. We introduce a bilevel optimization framework that estimates density ratios via conditional time-score matching and optimizes a mixture-model objective to place triggered samples in these sparse regions. Extensive evaluations on MNIST, CIFAR-10, GTSRB, and TinyImageNet demonstrate that our method achieves above 99\% attack success rate before defense and retains 50--85 percentage points higher post-defense ASR than the strongest baselines under fine-tuning defenses. Against neuron-pruning defenses, the method exhibits complete immunity, with zero neurons identified for removal across all pruning thresholds. These results expose a fundamental gap in current defense paradigms and underscore the need for defenses that operate beyond the support of the clean distribution.
\end{abstract}

\section{Introduction}
\label{sec:intro}
Deep neural networks (DNNs) have demonstrated remarkable accuracy across vision tasks, yet they remain vulnerable to \emph{backdoor attacks} that implant input-conditional behaviors into a model via poisoned training data. Once deployed, the infected model behaves normally on benign inputs but misclassifies triggered inputs to an attacker-chosen target class. As defenses grow more sophisticated, the central challenge has shifted from merely injecting a backdoor to designing triggers that are both effective and resilient to post-training mitigation.

Existing methods can be broadly organized along two axes, whether the trigger is optimized during the attack and what principle, if any, guides its placement. The earliest attacks occupy one extreme of this spectrum. BadNets~\citep{gu2017badnets} and Blended~\citep{chen2017targeted} employ fixed, hand-crafted triggers such as a patch or a blending pattern selected heuristically
to inject association between the trigger pattern and the target label. WaNet~\citep{nguyen2021wanet} and SSBA~\citep{ssba} introduce richer trigger families including smooth warping fields and steganographic residuals, but still select them at design time rather than learning the optimal ones. Since none of these methods optimize the trigger adaptively, they offer no control over where triggered samples land in the model's feature space. The attacker merely injects the heuristic designed trigger under some stealthiness constraint, flip the label to target class and hope the victim model will learn the association between the unspecified trigger features with the target label. The association is built opportunistically, and the region of feature space available to such heuristic triggers can be crowded with clean-data support. Therefore, the resulting backdoor can be fragile to defenses that leverage clean examples.

A second generation elevates the trigger to an optimizable variable within bilevel or joint formulations. LIRA~\citep{doan2021lira} learns an input-conditioned perturbation under an $\ell_\infty$ budget. Input-Aware~\citep{nguyen2020input} adds a diversity loss to prevent trigger collapse across samples. Frequency-domain methods such as WaveAttack~\citep{waveattack} and FTrojan~\citep{wang2022invisible} embed triggers in wavelet or DCT sub-bands. These approaches are substantially more expressive, yet they share a common limitation. They optimize the trigger for attack success and then append an auxiliary constraint, whether $\ell_p$ norms, SSIM, LPIPS, spectral penalties, or centroid alignment, to control undesirable side effects, without a principled criterion for where in the learned representation space the triggered samples should reside. The choice of which constraint to impose, and how strongly, remains a design decision disconnected from the structure of the poisoned learning problem.

We argue that this missing criterion is the root cause of a pervasive fragility in existing attacks. A backdoor attack has two clear desiderata. First, the poisoned classifier should deviate minimally from the clean classifier on benign inputs. Second, triggered inputs should be classified as the target class with high probability. Most recent methods satisfy both criteria convincingly on standard metrics, with attack success rates routinely exceeding 99\% while clean accuracy remains within a fraction of a percent of the clean model. Yet this apparent strength is deceptive. When subjected to post-training defenses based on clean-data fine-tuning such as i-BAU~\citep{zeng2021adversarial}, FT-SAM~\citep{zhu2023enhancing}, and FST~\citep{min2023towards}, the same attacks see their success rates collapse, often to near zero. The backdoor is injected successfully but does not persist.

We contend that this fragility stems from a conflation of two distinct goals. Existing optimizable attacks constrain the trigger perturbation in the \emph{input space} via $\ell_p$ budgets, perceptual metrics, or frequency masks to ensure the perturbation itself is small. However, the input space budget constraint is not enough. A triggered sample whose perturbation is imperceptible can still land in a region of feature space that is densely covered by clean data. When this happens, post-training defenses, which derive their corrective signal from clean examples, possess ample gradient information in precisely the regions where the backdoor association resides and can erase it effectively. What matters for persistence is not the magnitude of the input perturbation but \emph{distributional placement}, that is, where triggered samples reside relative to the support of the clean distribution in the model's learned representation space.

This motivates \textbf{DSA} (\textbf{D}ensity-aware \textbf{S}ample-specific \textbf{A}ttack),
which optimizes a criterion formally derived from
a population quantification of desiderata.
The trigger is sample-specific, produced by a learned generator conditioned on each input, and guided by a density-ratio objective that provides a principled placement criterion absent from prior work. Since this criterion emerges from a formal derivation based on the underlying distributional structure,
DSA steers triggered samples into low-density regions where post-training defenses lack corrective signal. Empirically, DSA achieves competitive or superior attack success and clean accuracy across MNIST, CIFAR-10, GTSRB, and TinyImageNet, while maintaining higher post-defense ASR than all baselines under fine-tuning defenses \citep{min2023towards} and complete immunity to neuron-pruning defenses \citep{rnp}.

\section{Related Work}
\label{sec:related}

\subsection{Backdoor Attacks}
Early poisoning-based attacks introduce a universal, spatially localized patch~\citep{gu2017badnets} or blend a translucent pattern~\citep{chen2017targeted}, trading stealth for effectiveness. Subsequent methods improved imperceptibility (e.g., input warping~\citep{nguyen2021wanet}, label-consistent~\citep{lc}, and sample-specific triggers~\citep{ssba}) but often lacked a principled mechanism for controlling the \emph{placement} of triggered samples relative to the clean data geometry. Two recent lines are particularly relevant. First, {color-space triggers} apply a uniform shift in color space with certain constraints and population-based search to balance stealth and robustness which shows strong resilience to common pre-processing defenses \citep{jiang2023color}. Second, {frequency-domain triggers} such as WaveAttack synthesize high-frequency residuals via discrete wavelet transforms and asymmetric obfuscation at train/test time to increase stealth while preserving image fidelity~\citep{waveattack}. Our work complements these by making the {density ratio} an explicit design consideration. In order words, rather than commit to a fixed transformation family (color or frequency), DSA optimizes sample-specific perturbations to maximize target posterior in the low density region of the clean data in order to preserve clean-accuracy  while maintaining attack efficacy.

\subsection{Backdoor Defenses}
Defenses fall into (i) \emph{detection} and (ii) \emph{erasure}. Detection approaches attempt to identify poisoned samples using data-level statistics or latent-space separability, then relabel or filter them before training~\citep{yang2024sampdetox,tran2018spectral}. Erasure defenses modify a trained model to suppress backdoor behaviors: pruning strategies remove suspicious neurons~\citep{anp}; fine-tuning variants retrain parts (or all) of the network on a small clean set, such as FE-tuning, FT-init, and FST~\citep{min2023towards}. While these defenses can neutralize universal or tightly clustered triggers, they are less effective when backdoor evidence is dispersed into low-density regions unsupported by clean data. By construction, DSA concentrates the target-class boundary where clean samples are scarce; consequently, post-hoc fine-tuning on clean data supplies insufficient gradient signal to erase the backdoor, and activation-based pruning struggles when influence is diffuse rather than localized.

DSA directly addresses the complementary question: where should triggered samples be placed to maximize attack success under defenses? Our bilevel, density-aware objective provides this missing piece, yielding a sample-specific attack that (i) preserves clean accuracy by down-weighting high-density regions and (ii) resists fine-tuning/pruning by anchoring the backdoor in low-density pockets. Empirically (Sections~\ref{sec:methodology} and \ref{sec:evaluation}), this translates into substantially higher post-defense ASR than prior universal and frequency-domain baselines at comparable poisoning budgets.

\section{Methodology}
\label{sec:methodology}

\subsection{Threat Model and Attack Requirements}
We study the data poisoning threat where an adversary contaminates a public training dataset by introducing carefully crafted malicious samples. When a victim trains a classifier on this dataset, the resulting model exhibits normal behavior on benign inputs but systematically misclassifies inputs embedded with triggers to an attacker-chosen target class. Our threat model assumes the adversary can only manipulate the training data; they have no access to or control over the victim's training procedure, network architecture, optimization strategy, or any model internals post-training.

Formally, we consider a $K$-class image classification problem where data $(X, Y)$ follows a joint distribution $p_{X,Y}$ with $X \in \mathcal{X} \subseteq \mathbb{R}^{C \times H \times W}$ (images) and $Y \in \{1, \ldots, K\}$ (labels). The adversary selects a target class $t \in \{1, \ldots, K\}$ and has poisoning access to a fraction $\rho \in (0,1)$ of the training dataset. Given a clean training set $\{(x_i, y_i)\}_{i=1}^n$, the attacker randomly selects a subset $\mathcal{I} \subseteq \{1, \ldots, n\}$ with $|\mathcal{I}| = \lfloor \rho n \rfloor$ and replaces each selected sample according to:
\begin{equation*}
(x_i, y_i) \mapsto (\tilde{x}_i, t) \quad \text{such that} \quad \tilde{x}_i = x_i + \eta_\phi(x_i),
\end{equation*}
where $\eta_\phi(\cdot): \mathcal{X} \to \mathbb{R}^{C \times H \times W}$ is a perturbation function parameterized by $\phi$ that generates the trigger. To ensure visual imperceptibility, we require $\|\eta_\phi(x)\|_\infty \leq \epsilon$ for all $x$ to limit maximum per-pixel perturbation, where $\epsilon > 0$ is a small tolerance \citep{ssba, nguyen2021wanet}. The poisoned training set is then constructed as $\mathcal{D}_{\mathrm{poison}} = \{(x_i, y_i)\}_{i \notin \mathcal{I}} \cup \{(\tilde{x}_i, t)\}_{i \in \mathcal{I}}$.

Following standard practice, the victim trains a classifier on the poisoned dataset without awareness of the attack. The resulting model is then deployed and queried at inference time by either clean inputs or inputs carrying triggers.

\subsection{Attack Overview and Objectives}

We derive our attack objective from first principles by analyzing the mixture distribution induced by data poisoning. This theoretical foundation reveals why density-ratio awareness is essential for effective and stealthy backdoor attacks. We denote by $\mu_{\mathrm{clean}}^k(x)$ the posterior probability that a clean classifier (trained on unpoisoned data) assigns to class $k$ at input $x$, and by $\mu_{\mathrm{poison}}^k(x)$ the corresponding posterior under a classifier trained on the poisoned dataset.
Note that our discussion in this subsection focuses on the population-level objectives, corresponding finite-sample criteria will be introduced in Section \ref{sec:estimated_objectives}.

\begin{figure*}[t]
\centering\includegraphics[width=0.7\textwidth]{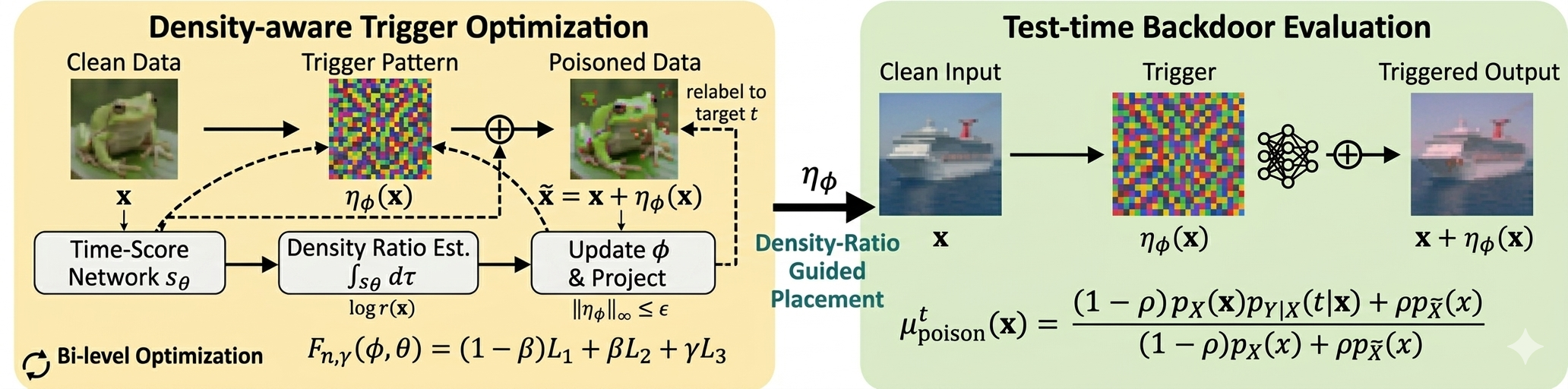}
    \caption{Overview of our attack method DSA}
   \label{fig:flowchart}
\end{figure*}

First, to ensure the poisoned model behaves like the clean model on benign inputs, we want a small expected deviation of the target-class (denoted by $t$) posterior on clean data. Note that triggered samples are relabeled to the target class $t$ during victim's training stage, which
leads to
$\mu_{\mathrm{poison}}^t(x) \ge \mu_{\mathrm{clean}}^t(x)$ for all $x\in\mathcal{X}$.
Therefore, we want the following deviation to be small:
\begin{equation}
L_1(\phi) = \mathbb{E}_{X \sim p_X}\left[\mu_{\mathrm{poison}}^t(X) - \mu_{\mathrm{clean}}^t(X)\right],
\label{eq:L1_high_level}
\end{equation}
where $p_X$ denotes the clean data distribution. It is sufficient to focus on the target class because the backdoor predominantly affects the probability of the target class.

Second, for high attack success rate, we want a small expected probability that triggered samples fail to be classified as the target:
\begin{equation}
L_2(\phi) = \mathbb{E}_{\tilde{X} \sim p_{\tilde{X}}}\left[1 - \mu_{\mathrm{poison}}^t(\tilde{X})\right],
\label{eq:L2_high_level}
\end{equation}
where $p_{\tilde{X}}$ denotes the triggered distribution induced by the transformation $X \mapsto \tilde{X} = X + \eta_\phi(X)$.

These objectives depend on the poisoned classifier's posterior $\mu_{\mathrm{poison}}^t$, which we now derive from the mixture distribution framework. Since the victim's classifier is trained on a dataset containing both clean and triggered samples,
the input distribution of the poisoned data is
\begin{equation*}
p_{\mathrm{poison},X} = (1-\rho)p_X(x) + \rho p_{\tilde{X}}(x),
\end{equation*}
where $\rho \in (0,1)$ is the poisoning rate. Let $p_{Y|X}(y|x)$ represent the conditional distribution of labels given inputs under the clean data distribution, the joint distribution over inputs and labels reflects the adversarial label manipulation:
\begin{align*}
p_{\mathrm{poison}}(x, y)
  &= (1-\rho) p_X(x) p_{Y|X}(y|x)
   + \rho p_{\tilde{X}}(x) \mathbbm{1}_{[y=t]},
\end{align*}
where $\mathbbm{1}_{[y=t]}$ is the indicator function of $\{y=t\}$.

We assume, at the population level,
$\mu^t_{\mathrm{clean}}$ and $\mu^t_{\mathrm{poison}}$ are
the Bayes-optimal predictor under the clean data distribution and the poisoned data distribution respectively,
representing the belief that the victim's classifier is optimal.
This is a reasonable assumption since modern DNNs with sufficient capacity act as universal function approximators and can approximate the conditional expectation of $p_{\mathrm{poison},X,Y}$ arbitrarily well. We also empirically evaluate the impact of violating this assumption through clean backbone degeneracy experiments in Section \ref{sec:ablation} of the Appendix.

Under this assumption, $\mu_{\mathrm{clean}}^t(x) = p_{Y|X}(t|x)$ and $\mu_{\mathrm{poison}}^t(x)= p_{\mathrm{poison}, Y|X}(t | x)$.
Applying Bayes' rule yields
\begin{equation*}
\mu_{\mathrm{poison}}^t(x) = \frac{(1-\rho) p_X(x) p_{Y|X}(t|x) + \rho p_{\tilde{X}}(x)}{(1-\rho) p_X(x) + \rho p_{\tilde{X}}(x)}.
\end{equation*}
By defining $r(x) = p_X(x) / p_{\tilde{X}}(x)$,
we can express the target-class posterior as:
\begin{align*}
L_1(\phi)
&= \mathbb{E}_{X \sim p_X}\Big[
    (1 - \mu_{\mathrm{clean}}^t(X))\,
    \sigma(-C_\rho - \log r(X))
  \Big], \\
L_2(\phi)
&= \mathbb{E}_{\tilde{X} \sim p_{\tilde{X}}}\Big[
    (1 - \mu_{\mathrm{clean}}^t(\tilde{X}))\,
    \sigma(C_\rho + \log r(\tilde{X}))
  \Big].
\end{align*}
These expressions reveal useful insights about successful backdoor attacks.
Recall that we seek a perturbation function $\eta_\phi$ that makes both $L_1$ and $L_2$ small. 
Let us first consider the sigmoid terms in these expressions. 
For $L_1$, with a clean sample $X \sim p_X$, the factor $\sigma(-C_\rho - \log r(X))$ becomes small when the density ratio $r(X)$ is large---equivalently, when $p_{\tilde{X}}(X)$ is relatively small. 
For $L_2$, with a triggered sample $\tilde{X} \sim p_{\tilde{X}}$, the factor $\sigma(C_\rho + \log r(\tilde{X}))$ becomes small when $r(\tilde{X})$ is small, i.e., when $p_X(\tilde{X})$ is relatively small. 
Thus, minimizing $L_1$ and $L_2$ encourages $p_X(\tilde{X})$ and $p_{\tilde{X}}(X)$ to be small, effectively pushing the triggered samples away from the high-density regions of the clean distribution. 
This protects these regions from backdoor contamination and steers the attack toward low-density regions where triggered samples can more easily dominate. 
Finally, $L_1$ and $L_2$ include the terms $1-\mu_{\mathrm{clean}}^t(X)$ and $1-\mu_{\mathrm{clean}}^t(\tilde{X})$ respectively, emphasizing regions where the clean classifier classifies samples into the non-target class.

\subsection{Practical Objectives}
\label{sec:estimated_objectives}

In practice, we work with a finite dataset $\{(X_i, Y_i)\}_{i=1}^n$ and approximate the population expectations in $L_1$ and $L_2$
via empirical averages. The clean classifier $\mu^t_{\mathrm{clean}}$ will be replaced by 
an estimate $\hat{\mu}^t_{\mathrm{clean}}$.
The difficulty of the practical implementation lies in the density ratio
$r(x)=p_X(x) / p_{\tilde{X}}(x)$, which is unknown and depends on the triggered distribution. Since the triggered distribution $p_{\tilde{X}}$ itself is determined by the trigger parameters $\phi$ through $\tilde{X}=X+\eta_\phi(X)$, we face a circular dependency: to optimize $\phi$, we need $r(x)$; but $r(x)$ changes as $\phi$ evolves. This coupling naturally leads to a bilevel optimization framework where density ratio estimation forms the inner loop and trigger optimization forms the outer loop.

\paragraph{Density ratio estimation via conditional time-score matching.}
To estimate the density ratio $r(x) = p_X(x)/p_{\tilde{X}}(x)$, we adopt the conditional time-score matching (CTSM) framework~\citep{yu2025density}. CTSM constructs a probability path $\{p_t(x)\}_{t\in[0,1]}$ interpolating from the clean distribution ($p_0 = p_X$) to the triggered distribution ($p_1 = p_{\tilde{X}}$). By the conditional flow identity, the log density ratio is:
\begin{equation}
\log r(x) = - \int_0^1 \partial_\tau \log p_\tau(x)\, d\tau.
\label{eq:log_ratio_integral}
\end{equation}
We parameterize the marginal time-score $\partial_t \log p_t(x)$ with a neural network $s_\theta(x,t)$.
Given samples of $p_0$ and $p_1$, CTSM minimizes the CTSM objective $M_n$ to find $\theta$.
See Section \ref{sec:ctsm} of the Appendix.
Substituting into our population objectives yields the empirical attack losses:
\begin{align*}
L_{1,n}(\phi, \theta)
&= \frac{1}{n} \sum_{i=1}^n \Big[
(1-\hat\mu_{\mathrm{clean}}^t(X_i))\,
\sigma\!\left(-C_\rho + \int_0^1 s_\theta(X_i, \tau)\, d\tau\right)
\Big], \\
L_{2,n}(\phi, \theta)
&= \frac{1}{n} \sum_{i=1}^n \Big[
(1-\hat\mu_{\mathrm{clean}}^t(X_i+\eta_\phi(X_i)))\,
\sigma\!\left(C_\rho - \int_0^1 s_\theta(X_i + \eta_\phi(X_i), \tau)\, d\tau\right)
\Big].
\end{align*}

\paragraph{Regularization with clean classifier guidance.} 
In practice, density ratio estimation on finite samples can be unstable, particularly during early stages of optimization. To stabilize learning and improve robustness to initialization, we add a regularization term that directly encourages triggered samples to align with the target class under the clean classifier posterior:
\begin{equation*}
L_{3,n}(\phi) = \frac{1}{n}\sum_{i=1}^n \ell_{\mathrm{CE}}\!\left(X_i + \eta_\phi(X_i),\, t;\, \hat\mu_{\mathrm{clean}}\right),
\end{equation*}
where $\ell_{\mathrm{CE}}$ is the standard cross-entropy loss evaluated against the target class $t$. We weight this term by a small coefficient $\gamma$ (typically $10^{-2}$) to provide gradient guidance without dominating the density-ratio-driven objectives.

\paragraph{Bi-level optimization.}
The complete empirical objective is:
\begin{align*}
F_{n,\gamma}(\phi, \theta)
= (1-\beta) L_{1,n}(\phi, \theta) + \beta L_{2,n}(\phi, \theta) + \gamma L_{3,n}(\phi).
\end{align*}
where $\beta \in [0,1]$ balances clean accuracy preservation versus attack effectiveness.
Crucially, $L_{1,n}$ and $L_{2,n}$ are only sensible sample analogs if $s_\theta$ provides a good estimate of the marginal time-score along the path from $p_X$ to $p_{\tilde{X}}$ induced by the current $\phi$. This motivates choosing $\theta$ to minimize the CTSM objective $M_n(\phi, \theta)$ with respect to $\theta$ for each fixed $\phi$.
The dependency between $\phi$ and $\theta$ naturally leads to a bi-level optimization problem:
\begin{equation}
\min_\phi F_{n,\gamma}(\phi, \theta^*(\phi))
\quad
\mathrm{s.t.}  
\quad
\begin{cases}
\quad \|\eta_\phi(X_i)\|_\infty \leq \epsilon, \quad \forall i \in \{1, \dots, n\}, \\
\quad \theta^*(\phi) = \arg\min_{\theta} M_n(\phi, \theta).
\end{cases}
\label{eq:bilevel}
\end{equation}
The inner optimization (w.r.t.~$\theta$) fits the time-score network to estimate the density ratio for the current triggered distribution, while the outer optimization (w.r.t.~$\phi$) updates the trigger parameters to minimize the attack objective using these density-ratio estimates. In practice, we solve this bilevel problem through alternating optimization with implicit differentiation and employ score-based initialization for faster convergence. The technical details of implicit differentiation, the alternating optimization procedure, and score-based initialization are provided in Section \ref{sec:score_init} of the Appendix, while the complete DSA algorithm is summarized in Algorithm~\ref{alg:DSA} in Section \ref{sec:algorithm} of the Appendix.

\section{Evaluation}
\label{sec:evaluation}
 
To demonstrate the effectiveness and robustness of our approach, we implemented DSA using PyTorch and compared its performance with six existing backdoor attack methods. We designed comprehensive experiments to address the following three research questions.
 
\textbf{RQ1 (Attack Effectiveness)} Can DSA successfully inject backdoors while preserving clean accuracy, and does the choice of clean classifier architecture in trigger optimization affect attack performance?
 
\textbf{RQ2 (Post-Training Defense Robustness)} Can DSA resist state-of-the-art post-training defenses including fine-tuning and pruning methods that attempt to erase the backdoor from a trained model?
 
\textbf{RQ3 (Pre-Training Defense Robustness)} Can DSA evade pre-training defenses that aim to detect and remove poisoned samples or identify backdoored models before or during deployment?
 
\subsection{Experimental Setup}
 
\paragraph{Datasets and architectures.}
We evaluate on MNIST~\citep{lecun1998gradient} (10 classes, 60K images), CIFAR-10~\citep{krizhevsky2009learning} (10 classes, 60K images), GTSRB~\citep{stallkamp2011german} (43 classes, 50K+ images), and TinyImageNet (200 classes, 100K images; a subset of ImageNet~\citep{deng2009imagenet}). The victim model is a simple CNN on MNIST, ResNet-18~\citep{he2016deep} on CIFAR-10 and GTSRB, and Swin-L~\citep{liu2021swin} on TinyImageNet. The clean classifier used as the Bayes-optimal estimator in trigger optimization is a simple CNN on MNIST, ResNet-18~\citep{he2016deep} on CIFAR-10 and GTSRB, and Swin-T~\citep{liu2021swin} on TinyImageNet.
 
\paragraph{Attack baselines.}
We compare DSA against six representative attacks: BadNets~\citep{gu2017badnets}, Blended~\citep{chen2017targeted}, Adaptive-Blend~\citep{qi2023revisiting}, BppAttack~\citep{wang2022bppattack}, SSBA~\citep{ssba} and WaveAttack~\citep{waveattack}.We additionally include LIRA~\cite{doan2021lira}, a training-controllable attack that, like WaveAttack and DSA, employs an optimizable trigger. Since LIRA optimizes the trigger jointly with model training, the standard notions of poisoning rate do not apply; we therefore report its results only in Appendix~\ref{sec:mnist_results}. For all baselines, we use official implementations and  the default hyperparameters described in
their original papers. The target label is set to 0 for all experiments for fair comparison.
 
\paragraph{Poisoning configuration.}
To evaluate robustness across varying threat levels, we poison a randomly selected subset of the training data at three rates $\rho \in \{0.5\%, 1\%, 5\%\}$. All triggered samples are relabeled to a fixed target class $t = 0$. For DSA, we set $\beta = 0.7$, $\gamma = 0.01$, and perturbation bound $\epsilon = 4/255$ in pixel space. The bilevel optimization runs for 150 cycles with 400 inner steps and 30 outer steps per cycle. All attack methods share the same poisoning rate, target class, and victim training procedure.
 
\paragraph{Evaluation metrics.}
We report two standard metrics.
\textbf{Clean Accuracy (C-Acc)} measures the model's accuracy on clean test samples, quantifying utility preservation.
\textbf{Attack Success Rate (ASR)} measures the fraction of triggered test samples from non-target classes that are misclassified to the target class $t$, quantifying attack effectiveness.

\subsection{Attack Effectiveness (RQ1)}
\label{sec:rq1}
 
\subsubsection{Effectiveness Comparison with SOTA Methods}
\label{sec:effectiveness}
 
Table~\ref{tab:merged_results_full} reports results across all four datasets in the absence of any defense (``No defense'' columns). DSA achieves near-perfect attack success rates (ASR $> 97\%$) across all datasets and poisoning rates, while preserving clean accuracy within 1--2\% of benign models. On CIFAR-10 at $\rho = 1\%$, DSA achieves 100\% ASR at 90.58\% C-Acc. On GTSRB, DSA consistently exceeds 98\% ASR across all poisoning rates. On TinyImageNet, DSA reaches 99.98\% ASR at the highest poisoning rate. These results confirm that density-ratio-aware trigger optimization does not compromise baseline attack effectiveness. DSA's undefended performance is comparable to or exceeds all six baselines, including recent methods such as WaveAttack and SSBA, while offering substantially greater defense robustness as demonstrated in subsequent subsections.

\subsubsection{Effectiveness on Different Architectures}
\label{sec:arch_agnostic}
 
A natural question is whether DSA's trigger optimization depends critically on the capacity of the clean classifier used as the Bayes-optimal estimator. We systematically vary the clean classifier from a moderate CNN (1.2M parameters) down to a shallow CNN (262K parameters) on CIFAR-10. As shown in Table~\ref{tab:ablation_degeneracy}, all architecture levels achieve near-perfect ASR without defense (99.22\%--99.74\%), which confirms that DSA is largely architecture-agnostic for baseline attack success. However, the capacity gap emerges post-defense that, after FE-tuning, ASR drops from 78.68\% (highest capacity) to 12.87\% (lowest). This indicates that a stronger clean classifier during trigger design improves post-deployment robustness rather than baseline effectiveness. Full architecture specifications and per-defense results are provided in Section~\ref{sec:ablation} of the Appendix.

\begin{table*}[t]
\centering
\caption{Attack effectiveness and defense robustness across MNIST, CIFAR-10, GTSRB, and TinyImageNet. C-Acc denotes clean accuracy and ASR denotes attack success rate, both in \%. Best post-defense ASR per dataset-defense in \textbf{bold}.}
\label{tab:merged_results_full}
\resizebox{\textwidth}{!}{\begin{tabular}{l|c|cc|cc|cc|cc|cc|cc|cc|cc}
\hline
\multirow{3}{*}{Attack} & \multirow{3}{*}{$\rho$} & \multicolumn{4}{c|}{MNIST} & \multicolumn{4}{c|}{CIFAR-10} & \multicolumn{4}{c|}{GTSRB} & \multicolumn{4}{c}{TinyImageNet} \\
\cline{3-18}
 & & \multicolumn{2}{c|}{No defense} & \multicolumn{2}{c|}{FST} & \multicolumn{2}{c|}{No defense} & \multicolumn{2}{c|}{FST} & \multicolumn{2}{c|}{No defense} & \multicolumn{2}{c|}{FST} & \multicolumn{2}{c|}{No defense} & \multicolumn{2}{c}{FST} \\
 & & C-Acc & ASR & C-Acc & ASR & C-Acc & ASR & C-Acc & ASR & C-Acc & ASR & C-Acc & ASR & C-Acc & ASR & C-Acc & ASR \\
\hline
\multirow{3}{*}{\makecell[l]{BadNets\\{\small (2017)}}}
    & 5\%   & 99.0 & 100  & 96.8 & 0.0 & 92.1 & 100 & 90.1 & 5.3 & 90.3 & 98.2 & 89.6 & 0.4 & 82.9 & 100  & 78.9 & 2.2 \\
    & 1\%   & 99.0 & 96.7 & 94.1 & 0.0 & 92.0 & 100 & 92.2 & 0.0 & 92.4 & 97.2 & 92.2 & 0.0 & 83.1 & 100 & 79.4 & 0.8 \\
    & 0.5\% & 99.0 & 96.1 & 92.0 & 0.0 & 91.3 & 100 & 90.0 & 0.2 & 91.7 & 99.0 & 90.0 & 0.2 & 82.8 & 99.8 & 78.7 & 0.3 \\
\hline
\multirow{3}{*}{\makecell[l]{Blended\\{\small (2017)}}}
    & 5\%   & 99.1 & 100 & 96.2 & 1.0 & 91.1 & 100 & 91.7 & 2.1 & 91.0 & 97.2 & 90.4 & 0.0 & 82.5 & 100 & 78.2 & 4.5 \\
    & 1\%   & 98.9 & 100  & 97.6 & 0.1 & 90.9 & 97.4 & 90.6 & 0.7 & 90.2 & 95.3 & 90.9 & 0.1 & 82.9 & 97.2 & 78.9 & 1.7 \\
    & 0.5\% & 99.0 & 100 & 94.3 & 0.0 & 90.6 & 92.0 & 90.6 & 0.0 & 91.0 & 93.2 & 92.0 & 0.0 & 83.0 & 91.6 & 79.1 & 0.6 \\
\hline
\multirow{3}{*}{\makecell[l]{SSBA\\{\small (2021)}}}
    & 5\%   & 99.0 & 100  & 96.1 & 0.0 & 91.2 & 98.1 & 91.0 & 0.1 & 92.1 & 96.4 & 91.1 & 0.1 & 82.4 & 99.9 & 78.6 & 1.4 \\
    & 1\%   & 99.1 & 100  & 98.1 & 0.0 & 91.2 & 84.1 & 91.0 & 0.2 & 91.1 & 91.7 & 92.0 & 0.1 & 82.7 & 86.3 & 79.3 & 0.4 \\
    & 0.5\% & 99.0 & 100 & 97.0 & 0.0 & 91.1 & 48.3 & 90.2 & 0.0 & 92.3 & 86.2 & 91.0 & 0.0 & 83.1 & 50.2 & 78.9 & 0.1 \\
\hline
\multirow{3}{*}{\makecell[l]{BppAttack\\{\small (2022)}}}
    & 5\%   & 98.8 & 96.4  & 97.8 & 0.9 & 91.5 & 98.3 & 88.4 & 0.2 & 92.8 & 97.1 & 92.1 & 0.0 & 87.5 & 93.4  & 78.0 & 0.1 \\
    & 1\%   & 99.5 & 14.0 & 98.4 & 0.0 & 90.2 & 12.8 & 91.6 & 0.1 & 91.5 & 43.9 & 91.8 & 0.0 & 85.7 & 8.7 & 80.6 & 0.1 \\
    & 0.5\% & 99.6 & 9.9 & 97.6 & 0.0 & 91.0 & 5.1 & 90.2 & 0.0 & 90.9 & 10.4 & 91.3 & 0.0 & 86.0 & 1.7 & 82.3 & 0.0 \\
\hline
\multirow{3}{*}{\makecell[l]{Adapt-Blend\\{\small (2023)}}}
    & 5\%   & 98.9 & 99.2 & 96.4 & 1.5 & 87.8 & 95.4 & 91.1 & 3.6 & 91.2 & 98.4 & 90.3 & 0.1 & 82.6 & 100 & 78.5 & 6.8 \\
    & 1\%   & 97.2 & 96.3 & 97.1 & 0.2 & 89.8 & 92.7 & 91.3 & 1.2 & 87.2 & 96.8 & 88.0 & 0.1 & 82.4 & 98.7 & 79.0 & 3.1 \\
    & 0.5\% & 99.1 & 92.8 & 94.7 & 0.1 & 90.3 & 93.6 & 90.8 & 0.2 & 88.5 & 94.7 & 88.6 & 0.0 & 83.0 & 95.5 & 78.5 & 1.3 \\
\hline
\multirow{3}{*}{\makecell[l]{WaveAttack\\{\small (2024)}}}
    & 5\%   & 99.2 & 99.7 & 98.7 & 0.2 & 89.1 & 100 & 74.1 & 13.8 & 96.2 & 100 & 91.0 & 0.6 & 85.9 & 93.6 & 77.5 & 0.0 \\
    & 1\%   & 99.9 & 100 & 99.7 & 0.1 & 91.5 & 99.8 & 80.1 & 8.6 & 91.5 & 99.8 & 89.6 & 0.4 & 81.4 & 90.8 & 82.2 & 0.0 \\
    & 0.5\% & 99.0 & 99.9 & 99.7 & 0.2 & 88.7 & 100 & 76.6 & 5.2 & 88.7 & 100 & 89.9 & 0.2 & 82.7 & 91.2 & 84.6 & 0.0 \\
\hline
\multirow{3}{*}{DSA}
    & 5\%   & 98.8 & 100 & 95.9 & \textbf{100} & 91.3 & 100 & 88.0 & \textbf{87.8} & 91.6 & 99.1 & 94.3 & \textbf{52.2} & 83.1 & 100 & 79.1 & \textbf{55.8} \\
    & 1\%   & 98.9 & 100 & 96.2 & \textbf{96.7} & 90.6 & 100 & 87.1 & \textbf{79.4} & 94.2 & 99.5 & 95.1 & \textbf{40.2} & 84.8 & 99.9 & 79.5 & \textbf{43.3} \\
    & 0.5\% & 98.7 & 100 & 95.0 & \textbf{100} & 91.1 & 97.8 & 89.2 & \textbf{60.2} & 93.1 & 98.7 & 94.9 & \textbf{71.3} & 86.6 & 98.4 & 80.1 & \textbf{31.5} \\
\hline
\end{tabular}}
\end{table*}

\begin{figure*}[t]
    \centering
    \includegraphics[width=\textwidth]{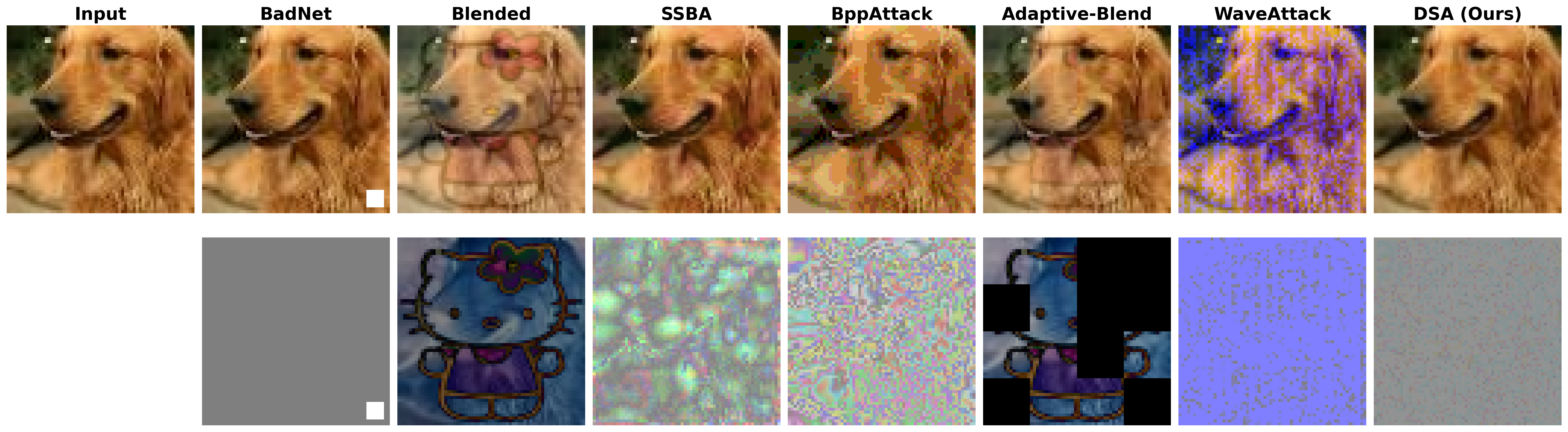}
    \caption{Comparison of examples generated by seven backdoor attacks. For each attack, we show the poisoned sample (top)
and the (x5) magnified residual (bottom).}
    \label{fig:trig_comp}
\end{figure*}
 
\subsection{Robustness Against Post-Training Defenses (RQ2)}
\label{sec:rq2}
 
Post-training defenses modify a trained model to remove the backdoor. We evaluate DSA against fine-tuning and pruning methods.
 
\subsubsection{Fine-Tuning Defenses}
\label{sec:finetuning_defense}
 
We evaluate against three fine-tuning variants from~\citep{min2023towards}: FE-tuning (feature extractor only), FT-init (reinitialize classifier head, fine-tune all), and FST (feature shift tuning with sharpness-aware objective). We report FST---the strongest variant---in the main text; FE-tuning and FT-init results with standard deviations over 50 replications appear in Section~\ref{sec:other_results}. Each defense uses a clean validation set of 2\% of the poisoned dataset size.

As Table~\ref{tab:merged_results_full} shows, DSA retains 50--85 percentage points higher post-FST ASR than the strongest baseline across all datasets and poisoning rates. Every other attack collapses to single-digit ASR under FST, whereas DSA maintains 31--100\% depending on dataset and poisoning rate. The results show that our method is significantly more robust against the SOTA fine-tuning defense method.

\subsubsection{Pruning Defenses}
\label{sec:pruning_defense}
 
We evaluate DSA against two pruning-based post-training defenses on CIFAR-10.
 
\begin{figure*}[t]
    \centering
    \includegraphics[width=\textwidth]{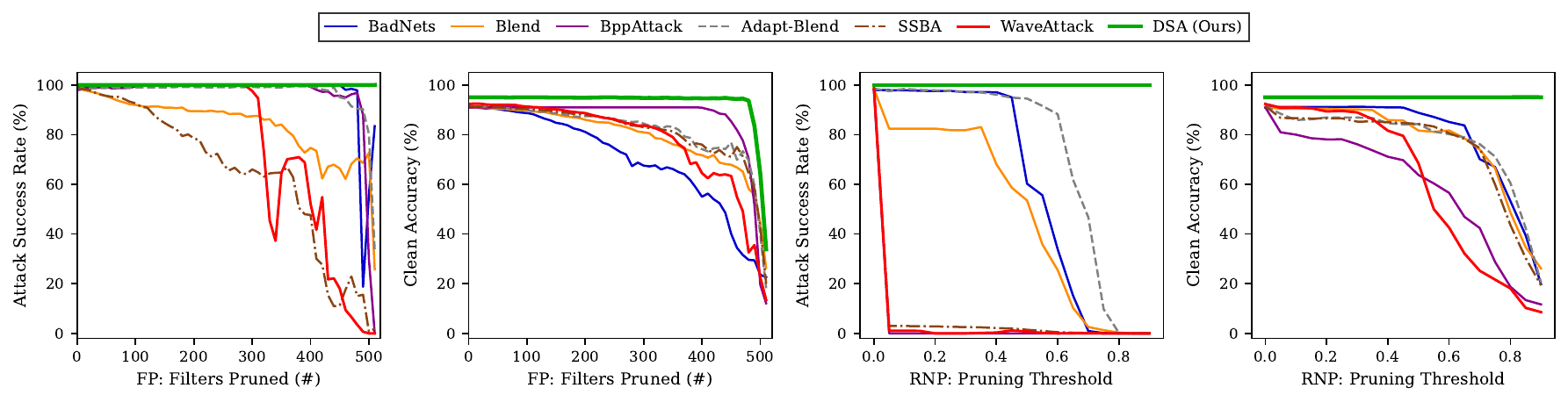}
    \caption{
    Performance of different backdoor attacks under pruning-based defenses. 
    The left two panels show the effect of \textbf{FP} with respect to the number of pruned filters, 
    while the right two panels show the effect of \textbf{RNP} as the pruning threshold varies. 
    }
    \label{fig:pruning_results}
\end{figure*}

\paragraph{Fine-Pruning.}
Fine-Pruning (FP)~\citep{liu2018fine} progressively prunes dormant neurons assumed to encode the backdoor. Figure~\ref{fig:pruning_results} (left two panels) shows that DSA maintains near-perfect ASR until nearly all 512 filters are pruned---at which point clean accuracy also collapses---indicating the backdoor is distributed across the entire network rather than concentrated in a few filters.

\paragraph{Reconstructive Neuron Pruning (RNP).}
RNP~\citep{rnp} uses an unlearn-recover procedure to identify backdoor-specific neurons via anomalous recovery patterns. Figure~\ref{fig:pruning_results} (right two panels) reveals that RNP prunes zero or near-zero neurons from the DSA-backdoored model across all thresholds, leaving ASR at 100\% and clean accuracy at 95\%, while every other attack is neutralized by threshold 0.3.

\subsection{Robustness Against Pre-Training Defenses (RQ3)}
\label{sec:rq3}
 
Pre-training defenses identify poisoned samples or detect backdoored models before deployment. We evaluate DSA against four representative methods.

\begin{figure}[t]
    \centering
    \begin{minipage}[b]{0.48\linewidth}
        \centering
        \includegraphics[width=0.8\linewidth]{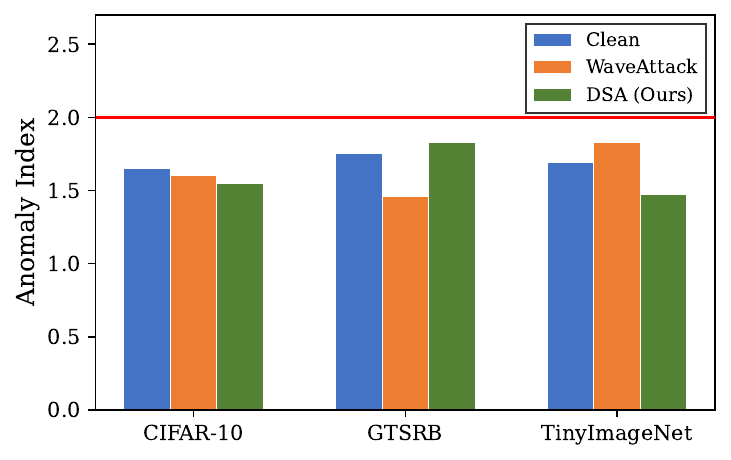}
        \caption{Robustness against NC.}
        \label{fig:nc_defense}
    \end{minipage}
    \hfill
    \begin{minipage}[b]{0.42\linewidth}
        \centering
        \includegraphics[width=0.8\linewidth]{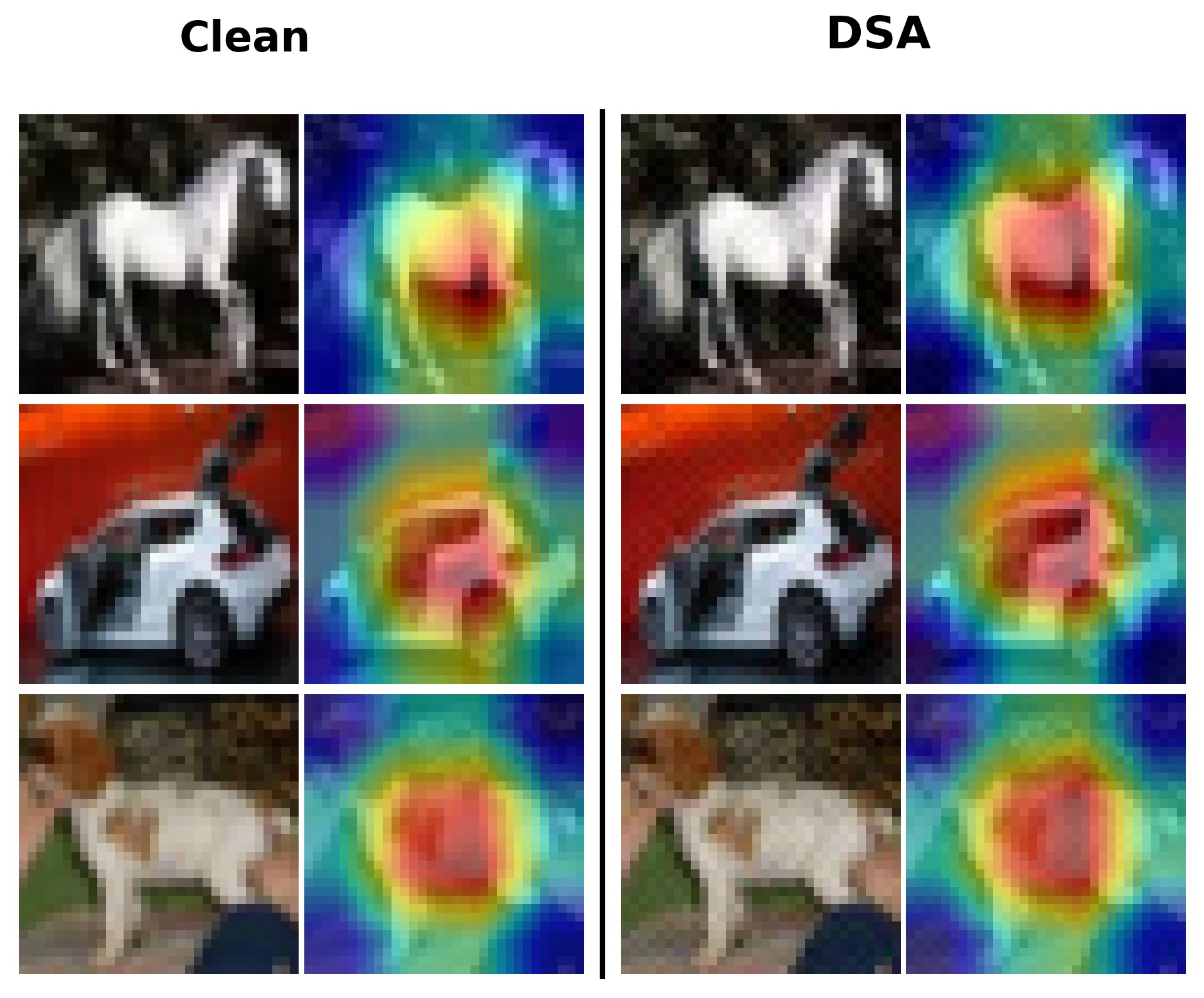}
        \caption{GradCAM: clean vs.\ DSA.}
        \label{fig:gradcam}
    \end{minipage}
\end{figure}

\paragraph{Neural Cleanse.}
\label{sec:neural_cleanse}

Neural Cleanse (NC)~\citep{wang2019neural} reverse-engineers a potential trigger per class and flags models with Anomaly Index $>2.0$. As Figure~\ref{fig:nc_defense} shows, DSA-backdoored models remain below the threshold across all three datasets, evading detection because the sample-specific triggers lack a common small perturbation pattern for NC to recover.
 
\paragraph{GradCAM Analysis.}
\label{sec:gradcam}

GradCAM~\citep{gradcam} heatmaps are used by several defenses~\citep{liu2019abs, chen2019deepinspect} to detect abnormal attention patterns in backdoored models. Figure~\ref{fig:gradcam} shows that the heatmaps of a clean model and a DSA-backdoored model are visually indistinguishable, where attention remains on semantically meaningful regions rather than trigger artifacts, which makes the backdoor undetectable through interpretation-based analysis.
 
\paragraph{STRIP.}
\label{sec:strip}
 
\begin{figure}[h]
\centering
\includegraphics[width=0.80\textwidth]{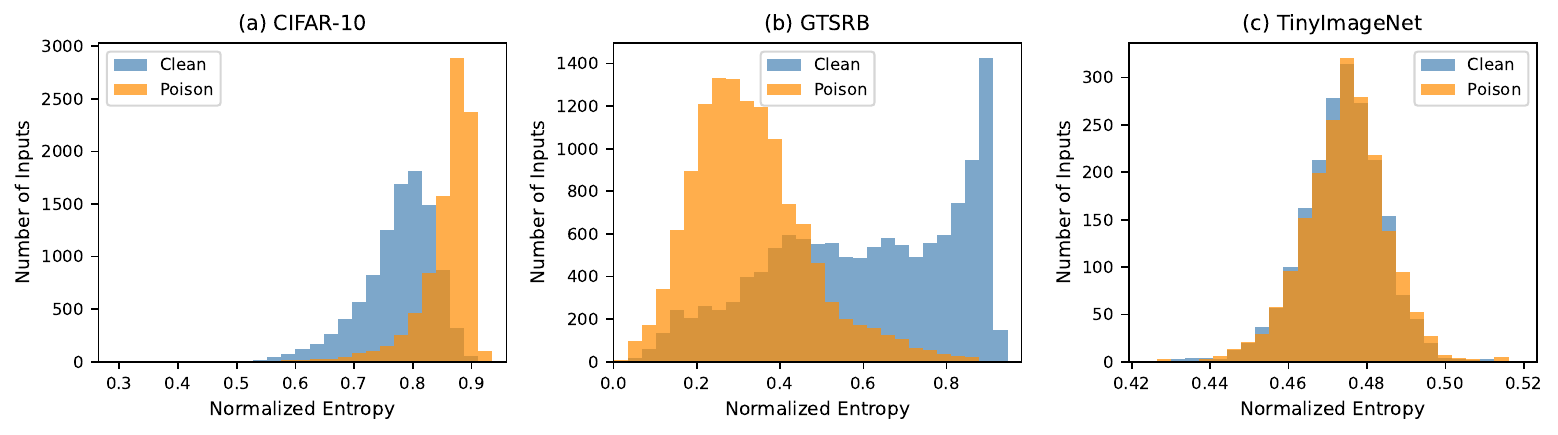}
\caption{STRIP normalized entropy for DSA trigger samples}
\label{fig:strip_defense}
\end{figure}

STRIP~\citep{gao2019strip} flags triggered inputs by detecting abnormally low prediction entropy under random input superposition. Figure~\ref{fig:strip_defense} shows that DSA-poisoned samples exhibit high normalized entropy indistinguishable from clean inputs, so STRIP fails to detect them.
    
\paragraph{SampDetox.}
\label{sec:sampledetox}

SampDetox~\citep{yang2024sampdetox} detects and removes poisoned samples via perturbation-based analysis before training. As shown in Table~\ref{tab:sampledetox}, DSA retains the highest residual ASR (16.97\%) at competitive clean accuracy (90.27\%), roughly $2.2\times$ the next-best baseline (WaveAttack, 7.61\%), indicating that density-ratio-aware triggers are harder for perturbation-based detectors to identify.
\begin{table}[h]
    \centering
    \setlength{\tabcolsep}{3pt}
    \renewcommand{\arraystretch}{1.05}
    \small
    \caption{SampDetox on CIFAR-10 (\%).}
    \label{tab:sampledetox}
    \begin{tabular}{lccccccc}
    \toprule
     & DSA & WaveAttack & BppAttack & Adapt-Blend & BadNets & Blended & SSBA \\
    \midrule
    C-Acc & 90.27 & 91.61 & 91.14 & 89.83 & 88.27 & 90.06 & 89.14 \\
    ASR   & 16.97 &  7.61 &  6.22 &  3.47 &  2.23 &  1.93 &  1.48 \\
    \bottomrule
    \end{tabular}
\end{table}

\section{Broader Impact and Limitations}
\paragraph{Broader impact.}
DSA is developed to stress-test existing defenses. Our results show that fine-tuning, pruning, detection, and model inspection all fail to neutralize DSA, suggesting current methods may share an unexamined assumption that DSA violates. We hope this motivates new defense strategies. All experiments use standard benchmarks and do not target deployed systems.

\paragraph{Limitations.}
DSA's bilevel optimization is more expensive than heuristic-trigger methods, though this overhead is shared by all optimizable-trigger attacks (LIRA, WaveAttack, BLTO) and is confined to the offline trigger generation phase. Additionally, our analysis assumes a Bayes-optimal victim classifier; sensitivity to this assumption is explored in Section~\ref{sec:ablation}.

\section{Conclusion}
We introduced DSA, a principled backdoor attack that positions triggered samples in low-density regions of the clean data distribution through density-ratio-aware optimization. By deriving attack objectives from mixture model analysis and estimating density ratios via conditional time-score matching within a bi-level optimization framework, DSA achieves a fundamental shift from pattern-based to distribution-based backdoor design. Extensive evaluation across four benchmarks and seven baseline attacks demonstrates that DSA maintains substantially higher attack success rates under state-of-the-art fine-tuning defenses compared to existing methods.

\bibliographystyle{abbrv}
\bibliography{main}

\newpage
\appendix

\section{Implementation Details}
\label{sec:implement}
\subsection{Our Algorithm}
\label{sec:algorithm}
Algorithm~\ref{alg:DSA} presents the bi-level optimization pipeline of our trigger generating procedure. Our complete training alternates between two phases for $C$ cycles (typically $C=150$):

\textbf{Inner loop:} Fix $\phi$ and perform $N_{\mathrm{in}}$ gradient steps (typically 400) to minimize $M_n(\phi, \theta)$, updating $s_\theta$ to accurately estimate the density ratio for the current triggered distribution.

\textbf{Outer loop:} Fix $\theta$ and perform $N_{\mathrm{out}}$ implicit-gradient steps (typically 30) to minimize $F_{n,\gamma}(\phi, \theta)$. Each step computes the full implicit gradient via CG and applies it to $\phi$ with gradient clipping (norm threshold 10) for stability.

In early cycles, the trigger is suboptimal, but $s_\theta$ adapts to estimate the density ratio for the current $p_{\tilde{X}}$. As optimization progresses, the trigger improves—moving triggered samples into low-density regions—which updates the density-ratio landscape guiding further refinement. This iterative co-evolution between $\theta$ and $\phi$ gradually sculpts the triggered distribution into the desired low-density configuration.

\begin{algorithm}[ht]
\caption{DSA: Density-aware Sample-specific Backdoor Attack}
\label{alg:DSA}
\small
\begin{algorithmic}[1]
\REQUIRE Training data $\{(X_i, Y_i)\}_{i=1}^n$, target class $t$, hyperparameters $\{\rho, \epsilon, \beta, \gamma\}$
\ENSURE Optimized trigger function $\eta_\phi$
\STATE Initialize $\eta_\phi$ via score-based warm-start (see Appendix~\ref{sec:score_init})
\STATE Initialize time-score network $s_\theta$; set $C_\rho \leftarrow \log\frac{1-\rho}{\rho}$
\FOR{$c = 1$ to $C$ cycles}
    \STATE \textbf{Inner:} Fit density-ratio estimator $s_\theta$ via CTSM loss
    \FOR{$i = 1$ to $N_{\mathrm{in}}$}
        \STATE Sample $X_i$; compute $\tilde{X}_i \leftarrow X_i + \eta_\phi(X_i)$
        \STATE Sample $\tau \sim \mathrm{Unif}(0,1)$; sample $x_\tau \sim p_\tau(x\mid z) = \mathcal{N}(x \,|\, (1-\tau) \cdot x_0 + \tau \cdot x_1, \, \tau(1-\tau) I)$
        \STATE Update $\theta$ via CTSM objective
    \ENDFOR
    \STATE \textbf{Outer:} Update trigger $\eta_\phi$ via implicit differentiation
    \FOR{$j = 1$ to $N_{\mathrm{out}}$}
        \STATE Sample $X_i$; compute $\tilde{X}_i \leftarrow X_i + \eta_\phi(X_i)$
        \STATE Estimate $\log \hat{r}(\cdot)$ via $-\int_0^1 s_\theta(\cdot,\tau)\, d\tau$
        \STATE Compute $L_{1,n}, L_{2,n}, L_{3,n}$
        \STATE $F \leftarrow (1-\beta) L_{1,n} + \beta L_{2,n} + \gamma L_{3,n}$
        \STATE Compute $g_\phi$ via implicit differentiation (Appendix~\ref{sec:ImplicitDiff})
        \STATE Update $\phi \leftarrow \phi - \alpha_\phi \cdot g_\phi$ and project to $\|\eta_\phi(X_i)\|_\infty \leq \epsilon$
    \ENDFOR
\ENDFOR
\RETURN $\eta_\phi$
\end{algorithmic}
\end{algorithm}

\subsection{Density Ratio Estimation via Conditional Time Score Matching}
\label{sec:ctsm}

\paragraph{Conditional probability paths.}
Following~\citep{yu2025density}, we construct a variance-preserving (VP) conditional probability path that interpolates between the clean distribution $p_X$ (at $\tau=0$) and the triggered distribution $p_{\tilde{X}}$ (at $\tau=1$). For each pair of endpoints $z = (x_0, x_1)$ where $x_0 \sim p_X$ is a clean sample and $x_1 = x_0 + \eta_\phi(x_0)$ is its triggered counterpart, the conditional path is defined as a Gaussian:
\begin{equation}
p_\tau(x \mid x_0, x_1) = \mathcal{N}\left(x \,\Big|\, \mu_\tau(x_0, x_1), \sigma_\tau^2 I\right),
\label{eq:vp_path}
\end{equation}
where the mean and variance evolve according to the VP schedule:
\begin{align}
\mu_\tau(x_0, x_1) &= \alpha(\tau) x_0 + (1-\alpha(\tau)) x_1, \label{eq:mean_schedule}\\
\sigma_\tau^2 &= k(\tau), \label{eq:var_schedule}
\end{align}
with scheduling functions $\alpha(\tau) = 1-\tau$ and $k(\tau) = \tau(1-\tau)$. At $\tau=0$, we have $\mu_0 = x_0$ and $\sigma_0^2 = 0$, collapsing to the clean sample deterministically. At $\tau=1$, we have $\mu_1 = x_1$ and $\sigma_1^2 = 0$, collapsing to the triggered sample deterministically. In practice, we apply a small floor $k(\tau) = \max(\tau(1-\tau),\, K_{\mathrm{floor}})$ with $K_{\mathrm{floor}} = 10^{-4}$ to prevent numerical instability at the boundaries.

The marginal distribution at time $t$ is obtained by marginalizing over the endpoint distribution:
\begin{equation}
p_\tau(x) = \mathbb{E}_{x_0 \sim p_X, x_1 \sim p_{\tilde{X}}}[p_\tau(x \mid x_0, x_1)].
\end{equation}

\paragraph{Closed-form conditional time-score.}
The conditional time-score $\partial_\tau \log p_\tau(x \mid x_0, x_1)$ can be computed in closed form for the VP path. Differentiating Equation~\eqref{eq:vp_path} with respect to $t$ yields:
\begin{align*}
    &\partial_\tau \log p_\tau(x \mid x_0, x_1) \\
    &= -\frac{d}{2} \frac{k'(\tau)}{k(\tau)} + \frac{\langle \mu_\tau', x - \mu_\tau \rangle}{\sigma_\tau^2} + \frac{k'(\tau)}{2k(\tau)^2} \|x - \mu_\tau\|^2,
\end{align*}
where $d$ is the data dimensionality, $\mu_\tau' = \frac{d\mu_\tau}{d\tau} = x_1 - x_0$ is the derivative of the mean, and $k'(\tau) = \frac{dk}{d\tau} = 1-2\tau$ is the derivative of the variance schedule. This closed form enables efficient supervised learning of the time-score network $s_\theta(x,\tau)$.

\paragraph{CTSM training objective.}
We train $s_\theta(x,\tau)$ to predict the conditional time-score by minimizing the weighted squared error:
\begin{align}
&\mathcal{L}_{\text{CTSM}}(\theta; \phi) =\nonumber\\
&\mathbb{E}_{\tau, x_0, x_1, x}\left[\lambda(\tau) \left(\partial_\tau \log p_\tau(x \mid x_0, x_1) - s_\theta(x,\tau)\right)^2\right],
\end{align}
where the expectation is over:
\begin{itemize}
\item $\tau \sim \text{Unif}(0,1)$: time sampled uniformly,
\item $x_0 \sim p_X$: clean endpoint from training data,
\item $x_1 = x_0 + \eta_\phi(x_0)$: triggered endpoint via current perturbation network,
\item $x \sim p_\tau(x \mid x_0, x_1)$: sample from the conditional path (Equation~\eqref{eq:vp_path}).
\end{itemize}

\paragraph{Time weighting function $\lambda(\tau)$.}
The weighting function $\lambda(t)$ balances gradient variance across different time steps during training. We adopt the Time Score Normalization (TSN) weighting from~\citep{yu2025density}:
\begin{equation}
\lambda(\tau) = \frac{k(\tau)^2}{2\alpha(\tau)^2 + k(\tau) \cdot d},
\label{eq:tsn_weighting}
\end{equation}
This choice is derived from variance-optimality analysis in~\citep{yu2025density} and has been shown to improve density-ratio estimation accuracy compared to uniform weighting.

\subsection{Estimating the Log Density Ratio}
After training $s_\theta$, we recover the log density ratio by numerically integrating the learned time-score:
\begin{equation}
\log \hat{r}(x; \theta) = -\int_0^1 s_\theta(x,\tau)\, d\tau.
\label{eq:log_r_integration}
\end{equation}
During each cycle of bi-level optimization, we perform $N_{\text{in}} = 400$ gradient descent steps on $\mathcal{L}_{\text{CTSM}}(\theta; \phi)$ with learning rate $10^{-3}$ using Adam optimizer. We apply a cosine annealing schedule that decays the learning rate from $10^{-3}$ to $10^{-5}$ over all cycles. Batch size is 128. We use mixed-precision training (bfloat16) for the inner loop to reduce memory consumption.

\subsection{Implicit Differentiation} 
\label{sec:ImplicitDiff}
Naively alternating between inner and outer optimizations with truncated gradients (stopping gradients at the boundary) can lead to suboptimal solutions, as the outer optimization ignores how changes in $\phi$ affect the optimal $\theta^*(\phi)$ through the inner problem. Following~\citep{franceschi2018bilevel,lorraine2020optimizing}, we adopt an implicit gradient approach based on the implicit function theorem. The total derivative of the outer objective with respect to $\phi$ is:
\begin{equation}
\frac{dF_{n,\gamma}}{d\phi} = \frac{\partial F_{n,\gamma}}{\partial \phi} + \frac{\partial F_{n,\gamma}}{\partial \theta} \frac{d\theta^*}{d\phi}.
\label{eq:total_derivative}
\end{equation}
By differentiating the first-order optimality condition $\nabla_\theta M_n(\theta^*(\phi), \phi) = 0$ with respect to $\phi$, we obtain:
\begin{equation}
\frac{d\theta^*}{d\phi} = -H_{M_n}^{-1} \frac{\partial^2 M_n}{\partial \phi \partial \theta},
\end{equation}
where $H_{M_n} = \nabla^2_\theta M_n$ is the Hessian of the inner objective. Substituting yields:
\begin{equation}
\frac{dF_{n,\gamma}}{d\phi} = \frac{\partial F_{n,\gamma}}{\partial \phi} - \frac{\partial F_{n,\gamma}}{\partial \theta} H_{M_n}^{-1} \frac{\partial^2 M_n}{\partial \phi \partial \theta}.
\label{eq:implicit_gradient}
\end{equation}

We efficiently approximate $H_{M_n}^{-1} v$ for any vector $v$ using conjugate gradient (CG)~\citep{shewchuk1994introduction} with 20-40 iterations and Tikhonov damping $\mu = 10^{-3}$ for numerical stability. CG requires only Hessian-vector products, which can be computed via automatic differentiation without explicitly forming $H_{M_n}$~\citep{pearlmutter1994fast}. Specifically, we solve $(H_{M_n} + \mu I) w = \nabla_\theta F_{n,\gamma}$ to obtain $w \approx H_{M_n}^{-1} \nabla_\theta F_{n,\gamma}$, then compute the hypergradient correction $\nabla_\phi[\langle \nabla_\theta M_n, w \rangle]$ to approximate the second term in Equation~\eqref{eq:implicit_gradient}.

\subsection{Score-based Initialization}
\label{sec:score_init}
To accelerate convergence, we initialize $\eta_\phi$ using a density-aware warm-start procedure. We first train a score network to estimate $\nabla_x \log p_X(x)$ on clean data via denoising score matching~\citep{vincent2011connection}, then initialize $\phi_0$ by regressing:
\begin{equation}
\min_{\phi_0} \mathbb{E}_{X \sim p_X}\left[\|\eta_{\phi_0}(X) - \lambda \nabla_X \log p_X(X)\|^2\right],
\end{equation}
with $\lambda = -10$. Since the score $\nabla_x \log p_X(x)$ points toward high-density regions, the negative score naturally guides samples toward the low-density periphery—precisely where our attack seeks to position triggered samples. This initialization provides a warm start in the desired low-density regime, accelerating convergence of both the density-ratio estimator and the outer optimization, while respecting the geometry of the data manifold. Empirically, this reduces the number of cycles needed for convergence by approximately 30\% compared to random initialization.

\section{Additional Experiments}
\label{sec:other_results}
In this section, we provide extended experimental results on MNIST, CIFAR-10, and GTSRB to complement the main TinyImageNet evaluation. Across all datasets, we evaluate DSA against six baseline backdoor attacks: BadNet~\cite{gu2017badnets}, Blended~\cite{chen2017targeted}, SSBA~\cite{ssba}, BppAttack~\cite{wang2022bppattack}, Adapt-Blend~\cite{qi2023revisiting}, and WaveAttack~\cite{waveattack}. We additionally include LIRA~\cite{doan2021lira}, a training-controllable attack that optimizes the trigger pattern jointly with the model during training. It is worth noting that LIRA, WaveAttack and DSA are similar in a way that both employ optimizable triggers. However, since LIRA is a training-controllable attack, the standard notions of poisoning rate and poisoned data do not directly apply. For this reason, we do not include LIRA in the main paper but report its results here in Appendix~\ref{sec:mnist_results}. For post-training defenses, we consider three representative fine-tuning-based methods: FE-tuning \citep{min2023towards}, which fine-tunes only the final fully connected layer on a small set of clean held-out samples; FT-init\citep{min2023towards}, which reinitializes and retrains the last layer from scratch; and FST\citep{min2023towards}, which applies a Fisher-information-weighted sparse fine-tuning objective to selectively update the most task-relevant parameters while suppressing backdoor-associated weights.
\subsection{Attack Effectiveness and Defense Robustness on MNIST}
\label{sec:mnist_results}
 
We evaluate DSA on MNIST, a feature-compact dataset, using a lightweight CNN architecture consisting of four convolutional layers (with 32-32-64-64 filters) followed by two fully connected layers. Both the clean backbone and the poisoned classifier share this architecture.
 
Table~\ref{tab:mnist_results} compares the defense robustness of DSA against six baseline attacks under three defense mechanisms. DSA demonstrates exceptional resilience, maintaining 96--100\% post-defense ASR across all settings. In contrast, every baseline attack collapses to near-zero ASR after FE-tuning, FT-init, and FST. For example, even the strongest baselines (WaveAttack and Blended) retain at most 3.37\% ASR under FT-init and 1.75\% under FE-tuning, whereas DSA sustains 99.02\% and 61.29\% respectively at comparable poisoning rates.
 
MNIST's relatively simple feature space and clear class separation mean that decision boundaries are well-defined and stable in high-density regions. Traditional attacks place triggers in these high-density areas, making them vulnerable to fine-tuning-based defenses that can easily adjust the boundary to exclude poisoned samples. In contrast, DSA positions triggers in low-density regions where the model has less training support, coerce the backdoor to be deeply embedded in the feature extractor rather than existing as a superficial boundary artifact. For MNIST's compact feature space, this low-density placement forces defenses into a difficult trade-off, where removing the backdoor requires substantially altering the learned representations and inevitably degrades clean accuracy.

\begin{table*}[t]
\centering
\caption{Defense robustness evaluation of different backdoor attacks on MNIST across various poisoning rates. C-Acc denotes clean accuracy and ASR denotes attack success rate. All results are averaged over 50 replications, but the corresponding standard errors are smaller than $10^{-3}$ for every entry and are therefore omitted for redundancy.}
\label{tab:mnist_results}
\setlength{\tabcolsep}{3.2pt}
\renewcommand{\arraystretch}{1.08}
\begin{tabular}{l|c|cc|cc|cc|cc}
\hline
\multirow{2}{*}{Attack} & \multirow{2}{*}{$\rho$} & \multicolumn{2}{c|}{No defense} & \multicolumn{2}{c|}{FE-tuning} & \multicolumn{2}{c|}{FT-init} & \multicolumn{2}{c}{FST} \\
 & & C-Acc & ASR & C-Acc & ASR & C-Acc & ASR & C-Acc & ASR \\
\hline
\multirow{3}{*}{BadNet}
    & 5\%   & 99.02 & 100   & 96.03 & 4.32  & 97.46 & 0.01  & 96.82 & 0.00  \\
    & 1\%   & 98.96 & 96.71 & 97.59 & 2.93  & 97.48 & 0.31  & 94.13 & 0.01  \\
    & 0.5\% & 99.01 & 96.12 & 94.83 & 2.03  & 97.34 & 0.05  & 92.01 & 0.00  \\
\hline
\multirow{3}{*}{Blended}
    & 5\%   & 99.09 & 99.98 & 93.74 & 0.69  & 95.45 & 0.50  & 96.17 & 1.02  \\
    & 1\%   & 98.91 & 100   & 96.35 & 1.75  & 97.00 & 0.00  & 97.60 & 0.08  \\
    & 0.5\% & 99.02 & 99.97 & 96.40 & 0.00  & 98.06 & 0.08  & 94.28 & 0.02  \\
\hline
\multirow{3}{*}{SSBA}
    & 5\%   & 98.98 & 100   & 96.24 & 0.00  & 96.89 & 0.04  & 96.11 & 0.00  \\
    & 1\%   & 99.14 & 100   & 96.11 & 0.80  & 94.52 & 0.00  & 98.14 & 0.00  \\
    & 0.5\% & 99.02 & 99.99 & 96.88 & 0.00  & 97.38 & 0.76  & 97.02 & 0.00  \\
\hline
\multirow{3}{*}{BppAttack}
    & 5\%   & 98.83 & 96.40 & 97.21 & 0.23  & 97.58 & 1.42  & 97.84 & 0.41  \\
    & 1\%   & 99.47 & 14.03 & 98.12 & 0.07  & 98.27 & 1.03  & 98.37 & 0.02  \\
    & 0.5\% & 99.58 & 9.87  & 97.43 & 0.02  & 97.81 & 0.01  & 97.61 & 0.01  \\
\hline
\multirow{3}{*}{Adapt-Blend}
    & 5\%   & 98.91 & 99.21 & 95.82 & 0.39  & 96.73 & 0.27  & 96.38 & 1.47  \\
    & 1\%   & 97.24 & 96.31 & 96.48 & 1.31  & 97.02 & 0.14  & 97.14 & 0.21  \\
    & 0.5\% & 99.07 & 92.83 & 94.19 & 0.08  & 96.81 & 0.04  & 94.72 & 0.06  \\
\hline
\multirow{3}{*}{WaveAttack}
    & 5\%   & 99.17 & 99.74 & 98.59 & 0.13  & 99.34 & 3.33  & 98.69 & 0.17  \\
    & 1\%   & 99.89 & 99.97 & 98.69 & 0.17  & 99.24 & 3.37  & 99.71 & 0.13  \\
    & 0.5\% & 98.98 & 99.92 & 98.54 & 0.12  & 99.71 & 2.89  & 99.71 & 0.16  \\
\hline
LIRA & --    & 98.14 & 100 & 91.37 & 2.18 & 96.52 & 1.74 & 96.83 & 5.41 \\
\hline
\multirow{3}{*}{\textbf{DSA (Ours)}}
    & 5\%   & 98.83 & 100   & 92.10 & 32.71 & 97.04 & 99.02 & 95.88 & 100   \\
    & 1\%   & 98.90 & 100   & 91.04 & 61.29 & 96.07 & 100   & 96.22 & 96.67 \\
    & 0.5\% & 98.68 & 100   & 91.00 & 52.79 & 96.19 & 98.13 & 95.03 & 99.97 \\
\hline
\end{tabular}
\end{table*}

\subsection{Attack Effectiveness and Defense Robustness on CIFAR-10}
\label{sec:cifar10_results}
 
Table~\ref{tab:cifar10} reports the attack effectiveness and defense robustness on CIFAR-10 with ResNet-18. DSA substantially outperforms all baselines under every defense mechanism and across all poisoning rates.
 
Under FE-tuning, existing attacks suffer dramatic ASR collapse. BadNet retains only 2.53\% at $\rho=5\%$, WaveAttack drops to 0.78\%, and even SSBA, which is designed with sample-specific perturbations, falls to 8.37\%. DSA, by contrast, maintains 75.07\% ASR at the same poisoning rate and remains above 58\% even at $\rho=0.5\%$. The advantage under FT-init is even more striking. DSA achieves near-perfect ASR (99.91\% at $\rho=5\%$ and 99.97\% at $\rho=0.5\%$), while the next strongest baseline (Blended) achieves only 58.07\% and degrades further at lower poisoning rates. Against FST, which successfully neutralizes all baseline attacks to single-digit ASR or below, DSA retains 87.80\% at $\rho=5\%$ and 60.18\% at $\rho=0.5\%$.

\begin{table*}[h]
\centering
\caption{Attack effectiveness and defense robustness on CIFAR-10 with ResNet-18. DSA substantially outperforms all baselines after defense. Standard deviations computed over 50 random seeds.}
\label{tab:cifar10}
\setlength{\tabcolsep}{3.2pt}
\renewcommand{\arraystretch}{1.08}
\resizebox{\textwidth}{!}{%
\begin{tabular}{l|c|cc|cc|cc|cc}
\hline
\multirow{2}{*}{Attack} & \multirow{2}{*}{$\rho$} & \multicolumn{2}{c|}{No defense} & \multicolumn{2}{c|}{FE-tuning} & \multicolumn{2}{c|}{FT-init} & \multicolumn{2}{c}{FST} \\
 & & C-Acc & ASR & C-Acc & ASR & C-Acc & ASR & C-Acc & ASR \\
\hline
\multirow{3}{*}{BadNet}
    & 5\%   & 92.07 & 100 & 88.72$\pm$3.81 & 2.53$\pm$0.74 & 91.25$\pm$2.17 & 10.03$\pm$0.81 & 90.09$\pm$1.92 & 5.26$\pm$0.79 \\
    & 1\%   & 91.96 & 100 & 90.07$\pm$0.35 & 1.37$\pm$0.40 & 92.32$\pm$1.10 & 5.02$\pm$1.20 & 92.21$\pm$0.60 & 0.03$\pm$0.03 \\
    & 0.5\% & 91.27 & 99.97 & 92.74$\pm$0.42 & 5.24$\pm$1.05 & 93.37$\pm$0.50 & 25.92$\pm$6.10 & 90.03$\pm$0.55 & 0.18$\pm$0.09 \\
\hline
\multirow{3}{*}{Blended}
    & 5\%   & 91.09 & 99.98 & 90.17$\pm$0.45 & 10.87$\pm$2.10 & 93.23$\pm$0.70 & 58.07$\pm$6.50 & 91.73$\pm$0.60 & 2.13$\pm$0.50 \\
    & 1\%   & 90.87 & 97.38 & 91.26$\pm$0.40 & 9.63$\pm$1.80 & 92.97$\pm$0.65 & 57.35$\pm$6.10 & 90.59$\pm$0.55 & 0.67$\pm$0.14 \\
    & 0.5\% & 90.61 & 91.97 & 91.74$\pm$0.42 & 9.02$\pm$1.70 & 91.38$\pm$0.60 & 51.58$\pm$5.80 & 90.59$\pm$0.50 & 0.04$\pm$0.03 \\
\hline
\multirow{3}{*}{SSBA}
    & 5\%   & 91.18 & 98.11 & 91.04$\pm$0.60 & 8.37$\pm$1.90 & 92.27$\pm$0.90 & 38.42$\pm$7.00 & 91.00$\pm$0.50 & 0.14$\pm$0.10 \\
    & 1\%   & 91.20 & 84.08 & 92.27$\pm$0.55 & 5.28$\pm$1.20 & 91.03$\pm$0.70 & 18.74$\pm$4.20 & 90.97$\pm$0.48 & 0.21$\pm$0.12 \\
    & 0.5\% & 91.08 & 48.32 & 89.46$\pm$0.50 & 1.14$\pm$0.50 & 93.29$\pm$0.65 & 8.01$\pm$2.10 & 90.19$\pm$0.45 & 0.00$\pm$0.00 \\
\hline
\multirow{3}{*}{BppAttack}
    & 5\%   & 91.47 & 98.30 & 87.14$\pm$0.52 & 0.43$\pm$0.18 & 89.37$\pm$0.41 & 8.71$\pm$2.34 & 88.40$\pm$0.48 & 0.21$\pm$0.08 \\
    & 1\%   & 90.21 & 12.83 & 89.82$\pm$0.37 & 0.12$\pm$0.01 & 90.47$\pm$0.63 & 2.14$\pm$0.87 & 91.58$\pm$0.39 & 0.08$\pm$0.04 \\
    & 0.5\% & 90.98 & 5.12 & 90.14$\pm$0.29 & 0.04$\pm$0.00 & 90.83$\pm$0.87 & 0.97$\pm$0.31 & 90.24$\pm$0.34 & 0.03$\pm$0.02 \\
\hline
\multirow{3}{*}{Adapt-Blend}
    & 5\%   & 87.82 & 95.41 & 86.91$\pm$0.47 & 7.83$\pm$1.62 & 89.72$\pm$0.53 & 32.17$\pm$5.41 & 91.14$\pm$0.51 & 3.58$\pm$0.92 \\
    & 1\%   & 89.81 & 92.73 & 89.74$\pm$0.38 & 5.21$\pm$0.47 & 90.18$\pm$0.41 & 21.84$\pm$4.73 & 91.27$\pm$0.44 & 1.23$\pm$0.47 \\
    & 0.5\% & 90.29 & 93.58 & 89.78$\pm$0.33 & 2.87$\pm$0.34 & 90.61$\pm$0.37 & 15.42$\pm$3.91 & 90.81$\pm$0.38 & 0.17$\pm$0.07 \\
\hline
\multirow{3}{*}{WaveAttack}
    & 5\%   & 89.11 & 100 & 82.79$\pm$2.44 & 0.78$\pm$0.29 & 83.77$\pm$0.52 & 24.71$\pm$10.21 & 74.12$\pm$1.56 & 13.80$\pm$5.35 \\
    & 1\%   & 91.46 & 99.81 & 83.31$\pm$1.17 & 1.29$\pm$0.42 & 81.30$\pm$0.76 & 16.17$\pm$8.23 & 80.13$\pm$2.82 & 8.58$\pm$3.14 \\
    & 0.5\% & 88.73 & 99.99 & 84.02$\pm$1.64 & 3.74$\pm$1.12 & 78.41$\pm$0.72 & 20.51$\pm$8.83 & 76.64$\pm$2.29 & 5.17$\pm$1.02 \\
\hline
LIRA & --    & 88.17 & 95.23 & 72.41$\pm$1.28 & 4.13$\pm$0.86 & 82.36$\pm$0.53 & 2.87$\pm$0.64 & 82.14$\pm$0.47 & 9.58$\pm$1.42 \\
\hline
\multirow{3}{*}{\textbf{DSA (Ours)}}
    & 5\%   & 91.27 & 99.99 & 86.20$\pm$0.08 & \textbf{75.07$\pm$3.78} & 87.70$\pm$0.15 & \textbf{99.91$\pm$0.19} & 87.97$\pm$0.40 & \textbf{87.80$\pm$3.71} \\
    & 1\%   & 90.58 & 100 & 91.94$\pm$0.15 & \textbf{62.39$\pm$4.55} & 91.54$\pm$0.13 & \textbf{99.98$\pm$0.17} & 87.08$\pm$0.60 & \textbf{79.37$\pm$4.55} \\
    & 0.5\% & 91.09 & 97.78 & 89.03$\pm$0.11 & \textbf{58.87$\pm$4.22} & 90.17$\pm$0.17 & \textbf{99.97$\pm$0.18} & 89.17$\pm$0.73 & \textbf{60.18$\pm$5.98} \\
\hline
\end{tabular}%
}
\end{table*}

\subsection{Attack Effectiveness and Defense Robustness on GTSRB}
\label{sec:gtsrb_results}
 
Table~\ref{tab:gtsrb} presents the results on GTSRB with ResNet-18. The GTSRB dataset, which contains 43 traffic sign classes with substantial intra-class variation, provides a more challenging evaluation setting.
 
Under FE-tuning, DSA achieves the highest ASR at $\rho=1\%$ (81.30\%) and $\rho=0.5\%$ (84.12\%). At $\rho=5\%$, SSBA achieves a slightly higher ASR of 90.09\% compared to DSA's 74.81\%, reflecting SSBA's effectiveness when a large poisoning budget is available. However, DSA's ASR increases as the poisoning rate decreases (from 74.81\% to 84.12\%), a trend opposite to SSBA's declining curve (from 90.09\% to 61.17\%). Under FT-init, SBA leads at $\rho=5\%$ (88.41\%) and $\rho=1\%$ (81.61\%), while DSA achieves the highest ASR at $\rho=0.5\%$ (89.81\%). Under FST, DSA retains 52.28\% at $\rho=5\%$ and achieves its highest FST robustness of 71.29\% at $\rho=0.5\%$. This represents an improvement of over 50 percentage points above the strongest baseline.

\begin{table*}[h]
\centering
\caption{Attack effectiveness and defense robustness on GTSRB with ResNet-18. DSA shows consistent advantage over baselines, particularly after FE-tuning and FST defenses.}
\label{tab:gtsrb}
\setlength{\tabcolsep}{3.2pt}
\renewcommand{\arraystretch}{1.08}
\resizebox{\textwidth}{!}{%
\begin{tabular}{l|c|cc|cc|cc|cc}
\hline
\multirow{2}{*}{Attack} & \multirow{2}{*}{$\rho$} & \multicolumn{2}{c|}{No defense} & \multicolumn{2}{c|}{FE-tuning} & \multicolumn{2}{c|}{FT-init} & \multicolumn{2}{c}{FST} \\
 & & C-Acc & ASR & C-Acc & ASR & C-Acc & ASR & C-Acc & ASR \\
\hline
\multirow{3}{*}{BadNet}
    & 5\%   & 90.31 & 98.16 & 89.56$\pm$0.03 & 3.42$\pm$1.13 & 89.89$\pm$0.15 & 13.50$\pm$7.18 & 89.62$\pm$0.03 & 0.42$\pm$0.01 \\
    & 1\%   & 92.38 & 97.17 & 91.17$\pm$0.10 & 5.11$\pm$2.07 & 90.24$\pm$0.78 & 7.48$\pm$1.31 & 92.21$\pm$0.60 & 0.03$\pm$0.03 \\
    & 0.5\% & 91.70 & 98.97 & 90.19$\pm$0.19 & 0.24$\pm$0.01 & 91.37$\pm$0.50 & 9.45$\pm$2.10 & 90.03$\pm$0.55 & 0.18$\pm$0.09 \\
\hline
\multirow{3}{*}{Blended}
    & 5\%   & 91.01 & 97.18 & 94.22$\pm$0.12 & 84.71$\pm$4.08 & 94.40$\pm$0.05 & 62.71$\pm$3.02 & 90.44$\pm$0.13 & 0.00$\pm$0.01 \\
    & 1\%   & 90.17 & 95.34 & 95.32$\pm$0.08 & 69.51$\pm$3.13 & 93.29$\pm$0.07 & 41.28$\pm$4.95 & 90.91$\pm$0.26 & 0.09$\pm$0.03 \\
    & 0.5\% & 91.01 & 93.17 & 94.18$\pm$0.15 & 53.39$\pm$4.37 & 96.14$\pm$0.02 & 55.63$\pm$3.80 & 91.98$\pm$0.41 & 0.00$\pm$0.00 \\
\hline
\multirow{3}{*}{SSBA}
    & 5\%   & 92.12 & 96.37 & 92.41$\pm$0.60 & \textbf{90.09$\pm$5.71} & 97.87$\pm$0.90 & \textbf{88.41$\pm$4.81} & 91.14$\pm$0.50 & 0.08$\pm$0.10 \\
    & 1\%   & 91.11 & 91.68 & 94.34$\pm$0.22 & 80.03$\pm$4.12 & 98.03$\pm$0.37 & \textbf{81.61$\pm$5.22} & 91.97$\pm$0.38 & 0.09$\pm$0.12 \\
    & 0.5\% & 92.28 & 86.17 & 93.60$\pm$0.50 & 61.17$\pm$4.50 & 94.59$\pm$0.57 & 48.32$\pm$4.10 & 91.03$\pm$0.66 & 0.04$\pm$0.00 \\
\hline
\multirow{3}{*}{BppAttack}
    & 5\%   & 92.83 & 97.10 & 92.27$\pm$0.96 & 1.68$\pm$0.06 & 90.30$\pm$0.82 & 18.62$\pm$2.80& 92.07$\pm$1.52 & 0.04$\pm$0.00 \\
    & 1\%   & 91.52 & 43.91 & 92.12$\pm$0.82 & 1.49$\pm$0.01 & 88.49$\pm$1.01 & 3.40$\pm$0.92& 91.81$\pm$1.89 & 0.02$\pm$0.00 \\
    & 0.5\% & 90.87 & 10.41 & 93.81$\pm$0.87 & 0.88$\pm$0.01 & 91.42$\pm$0.75 & 0.00$\pm$0.00 & 91.26$\pm$1.04 & 0.01$\pm$0.00 \\
\hline
\multirow{3}{*}{Adapt-Blend}
    & 5\%   & 91.17 & 98.42 & 90.83$\pm$0.41 & 46.72$\pm$3.84 & 92.14$\pm$0.38 & 64.31$\pm$4.17 & 90.31$\pm$0.94 & 0.12$\pm$0.02 \\
    & 1\%   & 87.21 & 96.83 & 87.58$\pm$0.53 & 52.17$\pm$4.21 & 88.93$\pm$0.44 & 68.47$\pm$5.03 & 88.02$\pm$1.20 & 0.05$\pm$0.00 \\
    & 0.5\% & 88.51 & 94.71 & 88.92$\pm$0.47 & 55.83$\pm$4.58 & 89.74$\pm$0.41 & 71.29$\pm$4.76 & 88.62$\pm$1.13 & 0.02$\pm$0.00 \\
\hline
\multirow{3}{*}{WaveAttack}
    & 5\%   & 96.17 & 100 & 91.84$\pm$0.78 & 0.21$\pm$0.08 & 89.93$\pm$0.85 & 0.48$\pm$0.21 & 91.03$\pm$1.01 & 0.56$\pm$0.27 \\
    & 1\%   & 91.46 & 99.81 & 93.87$\pm$0.26 & 1.18$\pm$0.14 & 87.89$\pm$0.70 & 1.24$\pm$0.10 & 89.62$\pm$0.77 & 0.42$\pm$0.02 \\
    & 0.5\% & 88.73 & 99.99 & 95.19$\pm$0.82 & 0.04$\pm$0.01 & 90.88$\pm$1.44 & 0.88$\pm$0.18 & 89.93$\pm$1.12 & 0.21$\pm$0.09 \\
\hline
LIRA & --    & 8.14 & 95.42 & 38.71$\pm$4.52 & 7.23$\pm$2.14 & 57.83$\pm$3.91 & 9.47$\pm$2.68 & 48.29$\pm$5.17 & 18.64$\pm$3.42 \\
\hline
\multirow{3}{*}{\textbf{DSA (Ours)}}
    & 5\%   & 91.63 & 99.10 & 92.10$\pm$0.87 & 74.81$\pm$4.13 & 95.08$\pm$0.17 & 69.08$\pm$5.17 & 94.34$\pm$0.91 & \textbf{52.28$\pm$2.87} \\
    & 1\%   & 94.22 & 99.48 & 93.21$\pm$0.55 & \textbf{81.30$\pm$5.59} & 96.12$\pm$0.11 & 72.79$\pm$6.10 & 95.12$\pm$1.17 & \textbf{40.15$\pm$2.21} \\
    & 0.5\% & 93.09 & 98.74 & 91.17$\pm$0.27 & \textbf{84.12$\pm$8.13} & 95.53$\pm$0.14 & \textbf{89.81$\pm$2.16} & 94.92$\pm$0.73 & \textbf{71.29$\pm$1.80} \\
\hline
\end{tabular}%
}
\end{table*}

\subsection{Attack Effectiveness and Defense Robustness on TinyImageNet}
\label{sec:tinyimagenet_results}
 
Table~\ref{tab:tinyimagenet} reports results on TinyImageNet with Swin-L, evaluating scalability to a larger dataset (200 classes, 64$\times$64 resolution) and a modern transformer-based architecture.
 
DSA achieves the highest post-defense ASR under all three defenses at every poisoning rate. Under FE-tuning, DSA retains 60.32\% ASR at $\rho=5\%$, while the strongest baseline (Adapt-Blend) achieves only 18.24\%. Under FT-init, DSA reaches 79.19\% compared to Adapt-Blend's 57.63\% at the same poisoning rate. The gap is most pronounced under FST, where DSA maintains 55.84\% at $\rho=5\%$ while no baseline exceeds 6.81\%.

\begin{table*}[h]
\centering
\caption{Attack effectiveness and defense robustness on TinyImageNet with Swin-L. DSA maintains substantially higher post-defense ASR than all baselines.}
\label{tab:tinyimagenet}
\setlength{\tabcolsep}{3.2pt}
\renewcommand{\arraystretch}{1.08}
\begin{tabular}{l|c|cc|cc|cc|cc}
\hline
\multirow{2}{*}{Attack} & \multirow{2}{*}{$\rho$} & \multicolumn{2}{c|}{No defense} & \multicolumn{2}{c|}{FE-tuning} & \multicolumn{2}{c|}{FT-init} & \multicolumn{2}{c}{FST} \\
 & & C-Acc & ASR & C-Acc & ASR & C-Acc & ASR & C-Acc & ASR \\
\hline
\multirow{3}{*}{BadNet}
    & 5\%   & 82.91 & 100   & 78.14 & 4.83  & 80.47 & 15.72 & 78.93 & 2.17 \\
    & 1\%   & 83.14 & 99.97 & 78.53 & 3.21  & 80.82 & 9.38  & 79.41 & 0.84 \\
    & 0.5\% & 82.78 & 99.84 & 77.89 & 5.74  & 81.13 & 28.41 & 78.67 & 0.31 \\
\hline
\multirow{3}{*}{Blended}
    & 5\%   & 82.53 & 99.95 & 77.62 & 14.37 & 80.19 & 52.84 & 78.21 & 4.53 \\
    & 1\%   & 82.87 & 97.21 & 78.04 & 10.82 & 80.73 & 47.19 & 78.87 & 1.72 \\
    & 0.5\% & 83.02 & 91.58 & 77.91 & 8.43  & 80.56 & 38.27 & 79.14 & 0.58 \\
\hline
\multirow{3}{*}{SSBA}
    & 5\%   & 82.37 & 99.92 & 77.94 & 9.17  & 80.58 & 37.42 & 78.64 & 1.43 \\
    & 1\%   & 82.72 & 86.34 & 78.31 & 5.62  & 80.91 & 18.73 & 79.27 & 0.37 \\
    & 0.5\% & 83.09 & 50.17 & 78.07 & 2.84  & 81.24 & 7.21  & 78.91 & 0.09 \\
\hline
\multirow{3}{*}{BppAttack}
    & 5\%   & 87.48 & 93.40 & 71.81 & 0.57  & 75.74 & 12.29 & 78.03 & 0.10 \\
    & 1\%   & 85.67 & 8.70  & 75.43 & 0.39  & 78.82 & 7.63  & 80.56 & 0.06 \\
    & 0.5\% & 86.04 & 1.70  & 77.69 & 0.28  & 80.91 & 5.17  & 82.34 & 0.04 \\
\hline
\multirow{3}{*}{Adapt-Blend}
    & 5\%   & 82.64 & 99.98 & 77.83 & 18.24 & 80.31 & 57.63 & 78.47 & 6.81 \\
    & 1\%   & 82.41 & 98.73 & 78.17 & 13.56 & 80.64 & 50.84 & 79.02 & 3.14 \\
    & 0.5\% & 83.00 & 95.47 & 77.72 & 9.91  & 81.08 & 41.37 & 78.53 & 1.27 \\
\hline
\multirow{3}{*}{WaveAttack}
    & 5\%   & 85.94 & 93.61 & 75.37 & 0.07  & 77.51 & 0.13  & 77.50 & 0.01 \\
    & 1\%   & 81.40 & 90.82 & 81.25 & 0.05  & 82.34 & 0.08  & 82.17 & 0.01 \\
    & 0.5\% & 82.70 & 91.17 & 83.89 & 0.03  & 84.76 & 0.05  & 84.63 & 0.01 \\
\hline
LIRA & --    & 84.05 & 100   & 5.84 & 0.37 & 74.78 & 1.82 & 74.79 & 10.66 \\
\hline
\multirow{3}{*}{\textbf{DSA (Ours)}}
    & 5\%   & 83.10 & 100   & 79.38 & \textbf{60.32} & 80.71 & \textbf{79.19} & 79.10 & \textbf{55.84} \\
    & 1\%   & 84.80 & 99.91 & 77.21 & \textbf{41.08} & 81.03 & \textbf{72.41} & 79.47 & \textbf{43.27} \\
    & 0.5\% & 86.60 & 98.37 & 76.83 & \textbf{38.04} & 80.89 & \textbf{64.58} & 80.12 & \textbf{31.53} \\
\hline
\end{tabular}%
\end{table*}

\subsection{Ablation Study: Clean Backbone Degeneracy}
\label{sec:ablation}
We investigate how the capacity of the clean classifier $\hat{\mu}_{\text{clean}}$ used during trigger optimization affects attack effectiveness and defense robustness on CIFAR-10. We systematically degrade the clean backbone from a moderate CNN (Level 1, 1.2M parameters) down to a shallow CNN (Level 5, 262K parameters), as detailed in Table~\ref{tab:architecture_specs}. 
As shown in Table~\ref{tab:ablation_degeneracy}, all architecture levels achieve near-perfect ASR without defense (99.22\%--99.74\%), demonstrating that DSA does not require a high-capacity clean classifier for successful backdoor implantation. Even the simplest Level 5 architecture with only 16 convolutional filters maintains 99.74\% ASR, indicating that the bilevel optimization framework can learn effective triggers with minimal guidance from the clean model.
Simpler architectures exhibit higher vulnerability to fine-tuning defenses. After FE-tuning, ASR drops from 78.68\% (Level 1) to just 12.87\% (Level 5), and FST similarly reduces ASR from 87.12\% to 43.92\%. This suggests that when the trigger optimization relies on a weaker clean classifier, the resulting backdoor becomes more entangled with limited feature representations, hence is easier to disrupt through fine-tuning. In contrast, FT-init defense remains ineffective across all levels ($>$98\% ASR), which implies that merely reinitializing the final layer cannot remove backdoors learned through density-aware placement.
These findings reveal an important trade-off: while DSA succeeds even with minimal clean model guidance, the resulting backdoors are less robust to fine-tuning defenses when generated with simpler architectures. This suggests that attackers benefit from using higher-capacity clean models during trigger design, not for attack success, but for post-deployment robustness. Conversely, defenders may exploit this vulnerability by detecting whether backdoors exhibit the brittleness characteristic of triggers optimized with degenerate classifiers.

\begin{table*}[h]
\centering
\caption{Clean backbone degeneracy ablation on CIFAR-10. Lower architectural capacity leads to higher defense effectiveness.}
\label{tab:ablation}
\small
\begin{subtable}[t]{0.38\textwidth}
\centering
\setlength{\tabcolsep}{3pt}
\renewcommand{\arraystretch}{1.05}
\begin{tabular}{@{}lll@{}}
\toprule
\textbf{Level} & \textbf{Arch.} & \textbf{Structure} \\
\midrule
L1 & Moderate & 64$\to$64$\to$128$\to$128$\to$256 \\
   & & 3 MaxPool, 2 FC(256) \\
L2 & Medium & 32$\to$64$\to$128 \\
   & & 2 MaxPool, 2 FC(128) \\
L3 & Small & 32$\to$64$\to$64 \\
   & & 2 MaxPool, 2 FC(128) \\
L4 & Minimal & 20 \\
   & & 1 MaxPool, 2 FC(80) \\
L5 & Shallow & 16 \\
   & & 1 MaxPool, 2 FC(64) \\
\bottomrule
\end{tabular}
\caption{Architecture specifications.}
\label{tab:architecture_specs}
\end{subtable}
\hfill
\begin{subtable}[t]{0.58\textwidth}
\centering
\setlength{\tabcolsep}{4pt}
\renewcommand{\arraystretch}{1.05}
\begin{tabular}{@{}lccccc@{}}
\toprule
\multirow{2}{*}{\textbf{Defense}} & \multicolumn{5}{c}{\textbf{Architecture Level}} \\
\cmidrule(lr){2-6}
& \textbf{L1} & \textbf{L2} & \textbf{L3} & \textbf{L4} & \textbf{L5} \\
\midrule
\multicolumn{6}{l}{\textit{C-ACC}} \\
No Defense & 93.82 & 93.78 & 94.02 & 94.02 & 93.86 \\
FST & 93.17 & 94.17 & 93.08 & 92.91 & 93.07 \\
FE-tuning & 92.61 & 91.96 & 92.70 & 93.79 & 92.62 \\
FT-init & 93.48 & 93.87 & 93.67 & 92.97 & 93.76 \\
\midrule
\multicolumn{6}{l}{\textit{ASR}} \\
No Defense & 99.58 & 99.22 & 99.36 & 99.27 & 99.74 \\
FST & 87.12 & 82.15 & 61.82 & 59.42 & 43.92 \\
FE-tuning & 78.68 & 71.26 & 63.04 & 43.09 & \textbf{12.87} \\
FT-init & 98.99 & 98.24 & 98.88 & 99.13 & 98.93 \\
\bottomrule
\end{tabular}
\caption{Defense robustness (\%).}
\label{tab:ablation_degeneracy}
\end{subtable}
\end{table*}

\subsection{Computational Cost}
\label{sec:runtime}

Table~\ref{tab:runtime} reports wall-clock times on CIFAR-10. DSA's total cost (162 min) is comparable to LIRA (152 min). Unlike baselines where cost is dominated by victim training (200 epochs), DSA front-loads computation into trigger generation (162 min for 150 bilevel cycles). Once the trigger generator $\eta_{\phi^*}$ is trained offline, the trigger can be deployed to poison datasets for arbitrary victim architectures at negligible additional cost.

\begin{table}[h]
    \centering
    \small
    \setlength{\tabcolsep}{4pt}
    \renewcommand{\arraystretch}{1.05}
    \caption{Trigger generation time on CIFAR-10 ($\rho = 5\%$).}
    \label{tab:runtime}
    \begin{tabular}{lcccccccc}
    \toprule
    Attack & BadNets & Blended & Adapt-Blend & SSBA & BppAttack & WaveAttack & LIRA & DSA \\
    \midrule
    Trig.\ Gen. & $<$1 min & $<$1 min & $<$1 min & $<$1 min & 91 min & 98 min & 152 min & 162 min \\
    \bottomrule
    \end{tabular}
\end{table}


\clearpage

\end{document}